\UseRawInputEncoding

\documentclass[letterpaper, 10 pt, conference]{IEEEtran}  

\IEEEoverridecommandlockouts                              

\usepackage{amsmath,amssymb,amsfonts}
\usepackage{algorithmic}
\usepackage{algorithm}
\usepackage{array}
\usepackage[caption=false,font=normalsize,labelfont=sf,textfont=sf]{subfig}
\usepackage{textcomp}
\usepackage{stfloats}
\usepackage{url}
\usepackage{verbatim}
\usepackage{graphicx}
\usepackage{cite}
\usepackage{color}
\usepackage{makecell}
\title{\LARGE \bf
Learning a Stable Dynamic System with a Lyapunov Energy Function for Demonstratives Using Neural Networks
}

\author{Yu Zhang$^{1}$, Yongxiang Zou$^{1}$, Haoyu Zhang$^{1}$, Xiuze Xia$^{1}$ and Long Cheng$^{1}$

\thanks{This work was supported in part by the National Natural Science Foundation of China under Grants 62025307, 62333023, and 62311530097.
	
	Y. Zhang, Y., Zou, H. Zhang,  X. Xia and L. Cheng, are all with the School of Artificial Intelligence, University of Chinese Academy of Sciences, Beijing 100049, China. They are also with the State Key laboratory of Multimodel Artificial Intelligence Systems, Institute of Automation, Chinese Academy of Sciences, Beijing 100190, China.  All correspondences should be addressed to the corresponding author Dr. Long Cheng (long.cheng@ia.ac.cn).}%
}

\begin{document}

\maketitle
\thispagestyle{empty}
\pagestyle{empty}
\columnsep 0.2in

\begin{abstract}

Autonomous Dynamic System (DS)-based algorithms hold a pivotal and foundational role in the field of Learning from Demonstration (LfD). Nevertheless, they confront the formidable challenge of striking a delicate balance between achieving precision in learning and ensuring the overall stability of the system. In response to this substantial challenge, this paper introduces a novel DS algorithm rooted in neural network technology. This algorithm not only possesses the capability to extract critical insights from demonstration data but also demonstrates the capacity to learn a candidate Lyapunov energy function that is consistent with the provided demonstrations. The model presented in this paper employs a simplistic neural network architecture that excels in fulfilling a dual objective: optimizing accuracy while simultaneously preserving global stability. To comprehensively evaluate the effectiveness of the proposed algorithm, rigorous assessments are conducted using the LASA dataset, further reinforced by empirical validation through a robotic experiment.
\end{abstract}

\section{INTRODUCTION}

The popularity of robotics is growing with advancements in technology, and robots are expected to possess greater intelligence to perform complex tasks comparable to those performed by humans. However, programming them, especially for non-experts, is challenging. In such situations, Learning from Demonstration (LfD) is a user-friendly approach for robots to acquire skills by observing demonstrations, without relying on an explicit script or defined reward functions \cite{b1,cr,my}.

In practical applications, achieving stability is crucial for ensuring that learned skills can effectively converge to the desired target state. Dynamic Systems (DS) represent powerful tools that offer a versatile solution for modeling trajectories and generating highly stable real-time motion, as highlighted in \cite{b2}. This superiority over traditional methods, such as interpolation techniques, underscores the value of DS in various contexts.

One notable advantage of Autonomous DS is their ability to encode the task's target state as a stable attractor, making them inherently resilient to perturbations. Consequently, DS can formulate stable motions from any starting point within the robot's workspace. This inherent stability facilitates seamless adaptation to new situations while maintaining robust performance. Furthermore, DS possess the capability to dynamically adjust the robot's trajectory to accommodate changes in the target position or unexpected obstacles, as discussed in reference \cite{b3}. This adaptability further enhances their utility in real-world scenarios.

Given the paramount importance of stability, ensuring the stability of DS is a critical consideration. One elegant approach to guaranteeing stability involves introducing a positive-definite and continuously differentiable Lyapunov function, which serves as a robust means of ensuring stability.
The first work that combines Lyapunov theorems with DS learning is known as the Stable Estimator of Dynamical Systems (SEDS) \cite{b2}. In SEDS, the objective is to learn a globally asymptotically stable DS, which is represented by Gaussian Mixture Models (GMM) and Gaussian Mixture Regression (GMR), while adhering to Lyapunov stability constraints.

However, it's worth noting that the utilization of a quadratic Lyapunov function in SEDS may introduce limitations on the accuracy of reproduction. Researchers have come to recognize that imposing overly strict stability constraints can curtail the accuracy of learning from demonstrations, leading to less precise reproductions \cite{b4}.

To address the problem, researchers have proposed several modified approaches aimed at enhancing reproduction accuracy while adhering to stability constraints. One such approach is the Control Lyapunov Function-based Dynamics Movements (CLF-DMs) algorithm, introduced in \cite{b5}. This algorithm operates in three steps: 
In the first step, the algorithm acquires a valid Lyapunov function that closely aligns with the available data. In the second step, state-of-the-art regression techniques are employed to model an approximation of the motion derived from the demonstrations. And the final step focuses on ensuring the stability of the reproduced motion in real-time by resolving a constrained convex optimization problem.

While CLF-DM has the advantage of being able to learn from a broader set of demonstrations, it is important to highlight that the algorithm requires solving optimization problems at each step is crucial to ensuring stability. This complexity and sensitivity to parameters are important considerations.

Another approach presented in reference \cite{b4} introduces a  transformation called "$\tau$ -SEDS." This transformation is intended to map the data onto a new  space, with the primary aim of improving accuracy while preserving the system's stability.

Compared to traditional methods, neural network-based algorithms \cite{b6,b7,b8,b9,b10} have proven to be highly effective for learning from demonstrations. These neural networks can be thought of as versatile fitting functions, and when appropriately structured or constrained, they can generate trajectories that converge to satisfy a Lyapunov function. This Lyapunov function can either be learned by another neural network or manually designed. In the pursuit of improving accuracy in reproduction, some researchers have proposed neural networks dedicated to estimating the Lyapunov function based on demonstration data, as demonstrated in \cite{b11}. This approach helps enhance reproduction accuracy by reducing violations of the Lyapunov function observed in the demonstration data. However, despite these efforts, the simplicity of the neural network structure in \cite{b11} may still leave room for improvement in reproduction accuracy.

To address this limitation, references \cite{b6,b7,b8,b9,b10} have introduced invertible neural network structures. These structures are used to transform the original DS into a new, simplified DS that possesses inherent properties that ensure convergence to the target. The advantage of using invertible transformations is that the fitted DS inherits stability from the simplified DS. However, it's important to note that invertible neural networks often have limitations in terms of their nonlinear fitting capabilities. To compensate for this, multiple layers of invertible neural networks may need to be stacked, which can increase computational costs and time.

To address the challenges commonly encountered by neural network-based algorithms, a novel neural network-based approach is introduced, which eliminates the necessity of invertible transformations. This paper's key contributions can be outlined as follows:
\begin{itemize}
	\item A novel neural network architecture is suggested, demonstrating the ability to directly learn a Lyapunov  function from provided data. 	
	\item The proposed neural network exhibits the ability to directly output state differentiations, aligning seamlessly with the requirements of DS. Leveraging the learned Lyapunov candidate function, the generated trajectories are guaranteed to converge to the desired target, offering enhanced stability and accuracy in the learning process.
	\item The experimental results conclusively show that the proposed algorithm possesses the versatility to not only learn effectively from a single demonstration but also learn from multiple demonstrations. This adaptability underscores the robustness and practicality of the proposed approach in a variety of learning scenarios. 
\end{itemize}

The paper is organized as follows:  Section \ref{sec2} provides a comprehensive discussion of the problem formulations. In section \ref{sec3}, the intricate details of the neural networks are presented. In section \ref{sec4}, the evaluation results of the proposed algorithm. This evaluation includes simulations conducted on various handwriting trajectories sourced from the Lasa datasets \cite{b12}, as well as experiments carried out on the Franka robot.  The conclusion is made in section \ref{sec5}.

\section{Problem Formulation}\label{sec2}
When a person or a robot is involved in a point-to-point task, the motions can frequently be efficiently modeled using an autonomous DS as:
\begin{equation}\label{equ:1}
	\dot{\boldsymbol{x}} = f(\boldsymbol{x}), \ \forall \boldsymbol{x} \in \mathbb{R}^{d},
\end{equation} where $f: \mathbb{R}^{d} \mapsto \mathbb{R}^{d}$ is a continuous and continuously differentiable function, which is described to have a single equilibrium state, denoted as $\boldsymbol{x}^{*}$. This equilibrium state represents the target state of the task and can be viewed as the attractor of the  DS.  To simplify the analysis and without sacrificing generality, the targets of the motions are assumed to be positioned at the origin of the Cartesian coordinate system by $\tilde{\boldsymbol{x}} = \boldsymbol{x} - \boldsymbol{x}^{*} $. Equation (\ref{equ:1}), which can also describe the manipulation skills, yields a solution $\Phi(\boldsymbol{x}_{0},t)$ that represents a valid trajectory generated when provided with  $\boldsymbol{x}_{0}$. The desired  motions can then be derived  by incorporating this trajectory with the target state $\boldsymbol{x} = \tilde{\boldsymbol{x}} + \boldsymbol{x}^{*}$. Consequently, by altering the initial conditions $\boldsymbol{x}_{0}$, different  trajectories leading to the target state $\boldsymbol{x}^{*}$ can be generated. This flexibility allows for various trajectories to be produced, each tailored to different initial conditions.
The demonstration data is usually structured in the following manner:
$\begin{Bmatrix}
	\boldsymbol{x}_{k,n},\dot{\boldsymbol{x}}_{k,n} \ | \ k = 1,2,...,K_{n}; \ n = 1,2,...,N_{d}
\end{Bmatrix}$
In this representation, $n$ denotes the index of the demonstrations, and $k$ represents the sampling time step. Here, $N_{d}$ indicates the total number of demonstrations, while $K_{n}$  signifies the total number of samples. The state variable $\boldsymbol{x}$ describes the state or configuration of either a human or a robot. The derivative $\dot{\boldsymbol{x}}$ denotes the first-order time derivative of $\boldsymbol{x}$. Additionally, it is crucial to emphasize that all demonstrations related to a specific task share the same target state.
 It is assumed that these demonstrations conform to an Autonomous DS as described in Equation (\ref{equ:1}). This alignment allows for the modeling of the system in a parametric fashion, which can be represented as:
\begin{equation}\label{equ:2}
	\hat{\dot{\boldsymbol{x}}} = \hat{f}(\boldsymbol{x},\boldsymbol{\theta}), \ \forall \boldsymbol{x} \in \mathbb{R}^{d}.
\end{equation}
where $\hat{\dot{\boldsymbol{x}}}$ represents the estimated value of $\dot{\boldsymbol{x}}$, and $\hat{f}$ signifies a manually designed model aimed at approximating the DS as depicted in (\ref{equ:1}), $\boldsymbol{\theta}$ denotes the parameters within the model that are subject to learning. The optimal parameter configuration, denoted as $\boldsymbol{\theta^{*}}$, is derived through the minimization of the following objective function:
\begin{equation}\label{equ:3}
	J(\boldsymbol{\theta} ) \propto \sum_{n=1}^{N_{d}}\sum_{k=1}^{K_{n}}\left \| \hat{\dot{\boldsymbol{x}}}_{k,n}-\dot{\boldsymbol{x}}_{k,n} \right \|^{2},
\end{equation}
where $\propto$ denotes the proportionality relation, $\left \| \cdot  \right \|^{2}$ denotes the $l_{2}$-norm and $\hat{\dot{\boldsymbol{x}}}_{k,n}$ means the estimate of $\dot{\boldsymbol{x}}_{k,n}$. 
Hence, the objective function ensures that the learned model produces accurate reproductions closely resembling the observed behavior.
Nonetheless, it's crucial to emphasize that the stability of the model cannot be guaranteed solely by the objective function. An Autonomous Dynamical System (DS) is deemed to be locally asymptotically stable at the point $\boldsymbol{x}^*$ if there exists a continuous and continuously differentiable Lyapunov candidate function $V(\boldsymbol{x}): \mathbb{R}^d\rightarrow \mathbb{R}$ that satisfies the following conditions:
\begin{equation}\label{equ:4}
	\begin{cases}
		(a)	V(\boldsymbol{x}^*)=0, \\
		(b)	\dot{V}(\boldsymbol{x}^*)=0, \\
		(c) V(\boldsymbol{x})>0 :\forall \boldsymbol{x}\neq \boldsymbol{x}^*, \\
		(d)	\dot{V}(\boldsymbol{x})<0 :\forall \boldsymbol{x}\neq \boldsymbol{x}^*.
	\end{cases}
\end{equation}
Additionally, if the radially unbounded condition as
\begin{equation}\label{equ:5}
	\lim_{ \left \| \boldsymbol{x} \right \|\rightarrow +\infty }V(\boldsymbol{x})=+\infty,
\end{equation} is met, the DS exhibits global asymptotic stability.

Integrating stability considerations into the DS learning process to guarantee the robot's achievement of the desired target state could potentially constrain the model's accuracy. This is because striving for accurate reproductions akin to the demonstration data might lead to violations of the pre-defined Lyapunov candidate function. However, deriving a suitable Lyapunov candidate function analytically from the demonstration data is a challenging task, rendering the search for a satisfactory solution quite difficult.

In this paper, a data-driven approach utilizing neural networks to learn the Lyapunov candidate function is proposed. This method offers an alternative that bypasses the need for designing Lyapunov candidate function, facilitating the achievement of stability in the DS learning process, which is introduced in the following section.

\section{The Proposed Approach}\label{sec3}
\begin{figure}[htbp]
	\centerline{\includegraphics[width=0.4\textwidth]{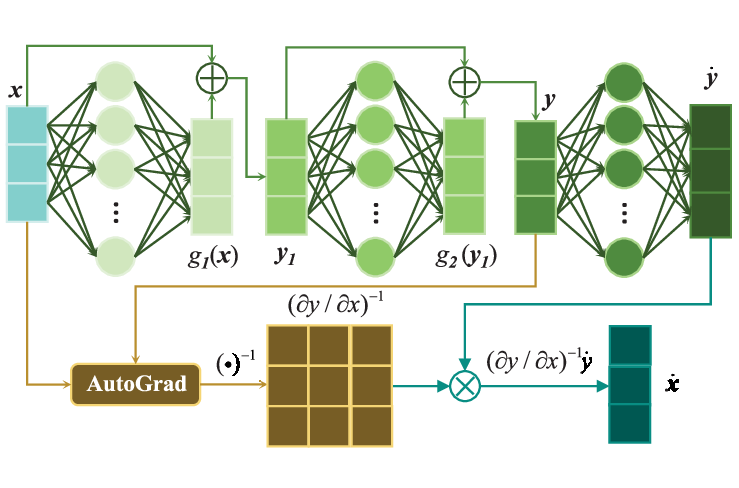}}
	\caption{The overall structure of the proposed algorithm for learning a DS while simultaneously learning a Lyapunov energy function.}	
	\label{fig1}
\end{figure}
To construct a Lyapunov energy function, the most straightforward way is to establish it as:
\begin{equation}\label{equ:6}
	V(\boldsymbol{x})=\frac{1}{2}\boldsymbol{x}^\mathrm{T}\boldsymbol{x} .
\end{equation}
Given the universal fitting capabilities of neural networks, the candidate Lyapunov energy function can be formulated as follows:
\begin{equation}\label{equ:7}
	V(\boldsymbol{x})=\frac{1}{2}g(\boldsymbol{x})^\mathrm{T}g(\boldsymbol{x}), 
\end{equation}
where $g$ is the function fitted by the neural networks.  Denoting $\boldsymbol{y}=g(\boldsymbol{x})$ as shown in Fig. \ref{fig1}, equation \ref{equ:7} becomes:
\begin{equation}\label{equ:ls}
	V(\boldsymbol{x})=\frac{1}{2}\boldsymbol{y}^\mathrm{T}\boldsymbol{y}. 
\end{equation}
 To ensure the stability of the system, it is essential to calculate the derivative of the energy function, which can be expressed as follows:
\begin{equation}\label{equ:8}
	\begin{aligned}	
	\dot{V}(\boldsymbol{x})&=\boldsymbol{y}^\mathrm{T} \dot{\boldsymbol{y}} \\
	&=\boldsymbol{y}^\mathrm{T}\frac{\partial \boldsymbol{y} }{\partial \boldsymbol{x}} \dot{\boldsymbol{x}} .
\end{aligned}
\end{equation}
When $\frac{\partial \boldsymbol{y} }{\partial \boldsymbol{x}}$ is invertible, $\dot{\boldsymbol{x}} $ can be calculated as: 
\begin{equation}\label{equ:9}
\dot{\boldsymbol{x}}=(\frac{\partial \boldsymbol{y} }{\partial \boldsymbol{x}})^{-1}\dot{\boldsymbol{y}},  
\end{equation}

By using the 
where $\dot{\boldsymbol{y}}$ is the designed variable in this paper.
Substituting the designed $\dot{\boldsymbol{x}}$  into \ref{equ:8}, the \ref{equ:8} becomes:

\begin{equation}\label{equ:10}
\begin{aligned}	
	\dot{V}(\boldsymbol{x})&=\boldsymbol{y}^\mathrm{T}\frac{\partial \boldsymbol{y} }{\partial \boldsymbol{x}} (\frac{\partial \boldsymbol{y} }{\partial \boldsymbol{x}})^{-1}\boldsymbol{y}
	 \\ &=\boldsymbol{y}^\mathrm{T}\dot{\boldsymbol{y}}
\end{aligned}
\end{equation}

 Without loss of generality, The motion targets are located at the origin of the Cartesian coordinate system, and the convergence point of $\boldsymbol{y}$ also coincides with the origin. As a result, when the following conditions are met, the global stability of the DS is ensured.
\begin{itemize}
	\item $\frac{\partial \boldsymbol{y} }{\partial \boldsymbol{x}}$ is invertible, which is the prerequisite condition of designing   $\dot{\boldsymbol{x}} $.	
	\item The appropriate $\dot{\boldsymbol{y}} $ to guarantee that $\boldsymbol{y}^\mathrm{T}\dot{\boldsymbol{y}}<0$, which is the required of the Lyapunov stability condition. 
	\item Only the target point of $\boldsymbol{x}^{*}$ should map to the convergence point of $\boldsymbol{y}$. In cases where the transformation is not invertible, and multiple points of $\boldsymbol{x}$ map to the convergence point of $\boldsymbol{y}$, then there is a risk that the original DS may converge to an undesired equilibrium point, which can result in unintended behavior or outcomes.
\end{itemize}

The first problem can be readily addressed by employing a residual structure for the design of $\boldsymbol{y}$, defined as:
\begin{equation}\label{equ:11}
	\boldsymbol{y} = m(\boldsymbol{x}) + \boldsymbol{x},
\end{equation}  where $m(\cdot)$ represents neural networks. In this context, the non-invertibility of $\frac{\partial \boldsymbol{y}}{\partial \boldsymbol{x}}$ occurs only when one of the eigenvalues of $\frac{\partial m(\boldsymbol{x})}{\partial \boldsymbol{x}}$ is equal to $-1$. However, this condition is highly unlikely when using neural network structures, making it a practical and effective solution.  
\begin{figure*}[htbp]
	\centerline{\includegraphics[width=1\textwidth]{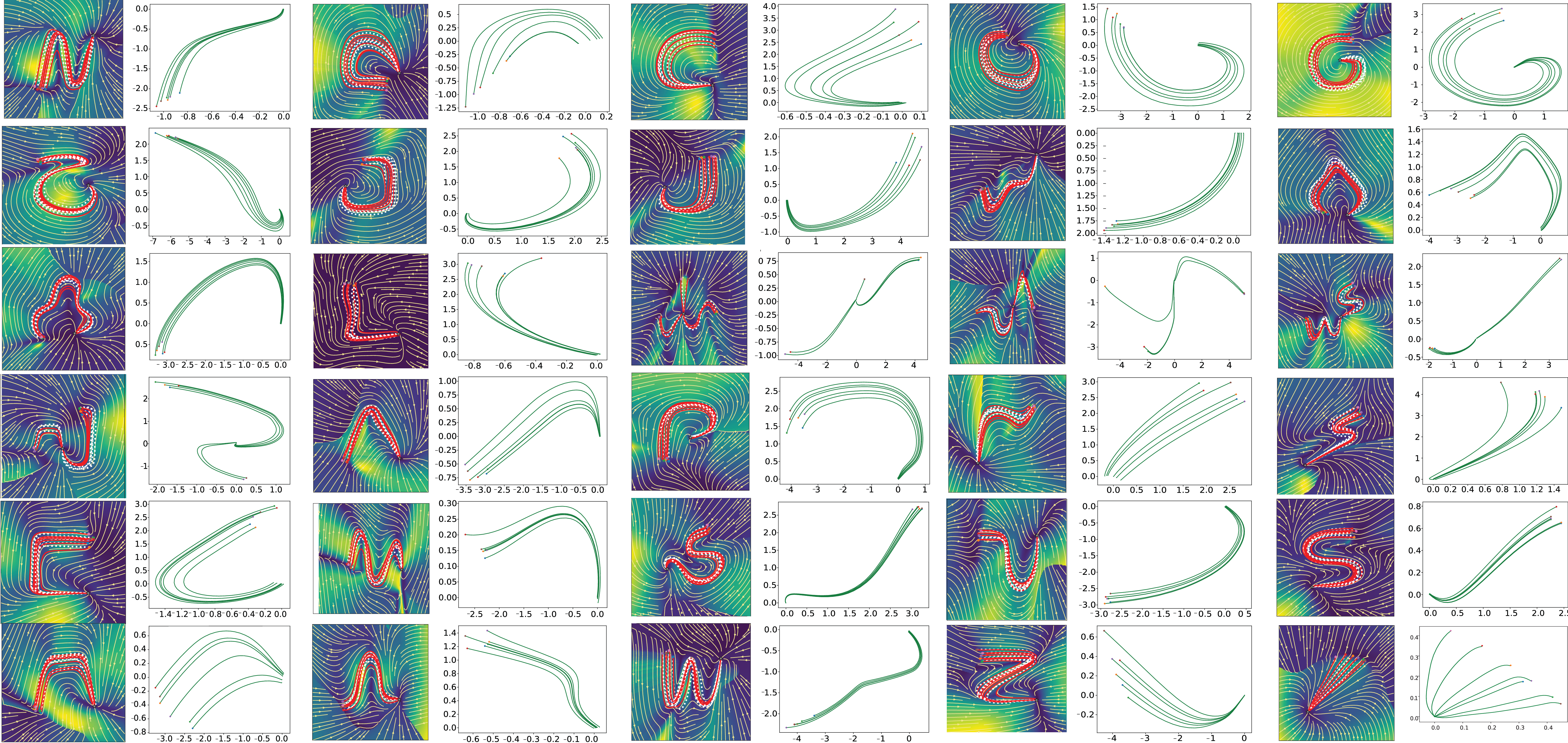}}
	\caption{
		The simulation utilizing the proposed algorithm is depicted in this paper. Images with a black-yellow background display the learned vector fields. Within these visuals, the white dotted lines represent the original demonstration data, while the red solid lines depict the reproductions from identical initial points. The target points are denoted as "$\cdot$" in these illustrations.	Moreover, the transformed reproduction trajectories (aligned with $\boldsymbol{y}$ as illustrated in Fig. \ref{fig1}) are showcased against a white background. Within this context, solid points indicate the starting positions for these trajectories.}	
	\label{fig2}
\end{figure*}
The second problem can be solved directly by designing $\dot{\boldsymbol{y}}$ as $\dot{\boldsymbol{y}}=-{\boldsymbol{y}}$.  An alternative method is to learn the $\dot{\boldsymbol{y}}$ using a neural network. To ensure compliance with the stability condition, it can be expressed as follows:
\begin{equation}\label{equ:12}
\boldsymbol{y}^\mathrm{T}\dot{\boldsymbol{y}} = 	\begin{cases}
	\boldsymbol{y}^\mathrm{T}	n(\boldsymbol{y}) \ \ \ \ \ \  if \ \boldsymbol{y}^\mathrm{T}	n(\boldsymbol{y})<-\beta \boldsymbol{y}^\mathrm{T}	\boldsymbol{y}, \\
	- \beta  \boldsymbol{y}^\mathrm{T}\boldsymbol{y} \ \ \ \ \ \ \ \ \	otherwise,
	\end{cases}
\end{equation}
where $n(\cdot)$ is a neural network and $\beta$ is a manually designed small positive constant. 
Taking inspiration from the research conducted by \cite{b13} and simultaneously learning both $\dot{\boldsymbol{y}}$ and $\dot{\boldsymbol{x}}$, the equation (\ref{equ:11}) can be reformulated as:
\begin{equation}\label{equ:13}
	\dot{\boldsymbol{y}} = 	
		n(\boldsymbol{y}) - \boldsymbol{y}  \frac{\mathrm{Relu}(\boldsymbol{y}^\mathrm{T}	n(\boldsymbol{y})+\beta \boldsymbol{y}^\mathrm{T}	\boldsymbol{y})}{\boldsymbol{y}^\mathrm{T}	\boldsymbol{y}}, 
\end{equation}
where $\mathrm{Relu}$ is a element-wise calculated function characterized by the following form:
\begin{equation}\label{equ:14}
	\mathrm{Relu}(s) = 	\begin{cases}
		s \ \ \ \ \ \ \ \ \  \mathrm{if} \ s>0, \\
		0 \ \ \ \ \ \ \ \ \	\mathrm{otherwise}.
	\end{cases}
\end{equation}

To address the third problem,  the neural network structure can be configured in a way that ensures the output becomes a zero vector only when the input is a zero vector. One simple way to enhance the performance of the neural network is to utilize a suitable activation function, such as the $\mathrm{Softplus}$  or $\mathrm{Sigmoid}$ function, which guarantee that the output is always greater than zero. Alternatively, the $\mathrm{Tanh}$ function can be employed to ensure that the output is always lesser than one. Furthermore, the output of the neural network should be multiplied with the input vector $\boldsymbol{x}$ and then added to it.
		
In our paper, the expression for $\boldsymbol{y}$ corresponding with the illustration in Fig. \ref{fig1} that is defined as follows:
\begin{equation}\label{equ:15}
	\begin{cases}
	g_{1}(\boldsymbol{x}) = m_{1}(\boldsymbol{x}),\\
	\boldsymbol{y}_{1} = g_{1}(\boldsymbol{x})+\boldsymbol{x},\\
	g_{2}(\boldsymbol{y}_{1}) = m_{2}(\boldsymbol{y}_{1})\odot \boldsymbol{y}_{1},\\
	\boldsymbol{y} =g_{2}(\boldsymbol{y}_{1})
	+\boldsymbol{y}_{1}
		\end{cases}	
\end{equation}
where $\odot$ denotes Hadamard product, the neural network $m_{1}(\cdot)$ in this paper utilizes the $\mathrm{PReLU}$  activation function and does not incorporate bias in any of its layers, the neural network $m_{2}(\cdot)$ employs the $\mathrm{Softplus}$ activation function and does not include bias in the final layer. Consequently, the proposed algorithm involves three neural networks: $m_{1}(\cdot)$, $m_{2}(\cdot)$, and $n(\cdot)$. The neural network $n(\cdot)$ employs the $\mathrm{Softplus}$ activation function, all the neural networks are configured with three layers in this paper.

 Interestingly, when there is a one-to-one correspondence between the target points in two spaces, the input is $\infty$ then the output is also $\infty$, and the two spaces have the same dimensionality, the transformation can be referred to as a bijective transformation.
By contradiction, the two-dimensional space is first taken into consideration, and it is assumed that the transformation from $\boldsymbol{x}$ to $\boldsymbol{y}$ is not bijective. In this case, it can be deduced that two vectors $\boldsymbol{x}$ are transformed to the same vector $\boldsymbol{y}$, such as $g(\boldsymbol{x_{1}})=g(\boldsymbol{x_{2}})$. Since equation \ref{equ:7} sets the Lyapunov energy function, it follows that the energy of both $x_1$ and $x_2$ is the same. In the transformed two-dimensional space, points that possess identical energy levels can be aggregated to constitute a single, closed, and continuous circle. Analogously, in the original space, points sharing the same energy level are expected to form two separate, continuous curves that intersect at points $x_1$ and $x_2$​, respectively. Utilizing the Lyapunov energy function, it is established that the energy of points situated on these closed curves must be less than that of the points
$x_1$ and $x_2$. Consequently, the existence of such closed curves becomes untenable, necessitating the conclusion that the two-dimensional transformation in question is indeed bijective. Furthermore, when the dimensionality of the transformation space exceeds two, a similar conclusion can be deduced by conceptualizing a closed, high-dimensional volume. Thus, it is justified to characterize the transformation as bijective across higher dimensions.

\section{ Experiment Results and Discussions}\label{sec4}
To comprehensively assess the efficacy of the proposed algorithm, a rigorous evaluation protocol encompassing both simulation and real-world experimentation has been implemented. The results are summarized as follows:

\subsection{ Simulation Results}
The initial assessment was carried out using the LASA dataset, which comprises a comprehensive collection of handwritten letter trajectories. 

To assess the effectiveness of the proposed algorithm, the quantitative evaluation of accuracy was performed through the application of two error metrics: the Swept Error Area (SEA) \cite{b5} and the Velocity Root Mean Square Error ($V_{rmse}$) \cite{b14}. The SEA scores serve as an indicator of the algorithm's capability to faithfully replicate the shapes of motions. On the other hand, the $V_{rmse}$ metric gauges the algorithm's proficiency in preserving the velocities inherent in demonstrations. A lower SEA signifies superior accuracy in reproducing the trajectories, while a lower $V_{rmse}$ indicates that the reproductions closely emulate the smoothness observed in the original demonstrations. These two metrics comprehensively evaluate the performance of the reproductions by quantifying their resemblance to the demonstrated trajectories.
\begin{figure}[htbp]	\centerline{\includegraphics[width=0.25\textwidth]{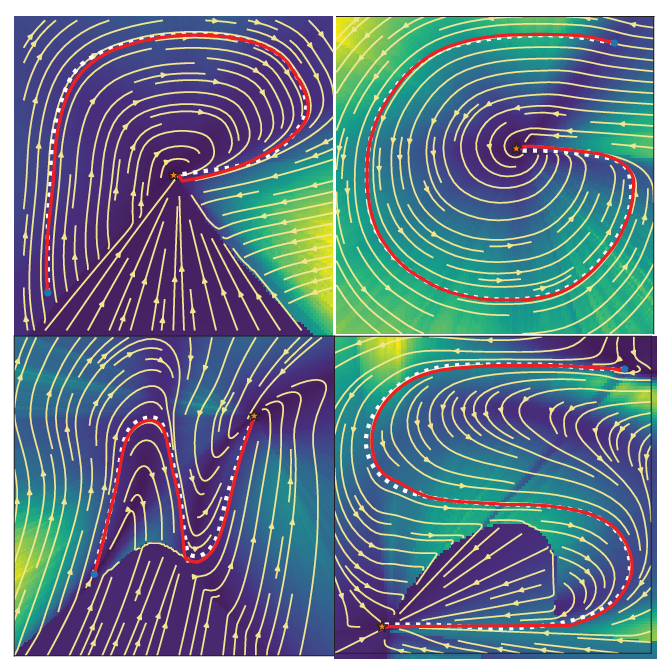}}
	\caption{Four simulation scenarios of learning from one demonstration by using the proposed algorithm are depicted. }	
	\label{fig3}
\end{figure}
For the implementation of the proposed algorithm, the "AdamW" optimization method is opted. Specifically, the learning rate was configured at $1 \times 10^{-5}$,   coupled with a decay rate of $0.99$. Subsequently, the trained DS model was employed to generate trajectories starting from identical points and with matching step increments as the original demonstrations. In this paper, all  data from the LASA dataset were utilized, with trajectories generated from all available starting points. Prior to processing, the data underwent normalization, bringing it within the range of [$-1, 1$]. The algorithm's parameters were initialized with random values, and the maximum number of iterations was set to 2000, with a mini-batch size of 64. For the sake of equitable comparisons,  DS algorithms that named SEDS and CLF-DM are chosen, for which source code was readily accessible. The parameters for these comparative algorithms were configured to match the values originally specified in their respective references, ensuring a fair benchmark.

Fig. \ref{fig2} presents a visual representation of the vector fields (depicted with dark background) and the transformed trajectories generated by the proposed DS algorithm for 29 examples drawn from the LASA dataset. Notably, the reproduced trajectories (depicted in red) closely align with the original demonstrations (depicted in white). It is worth emphasizing that  irrespective of whether the starting points align with those of the demonstrations, the reproduced DS consistently converges toward the intended goal. Furthermore, the last three images in the third row and first image in the fourth row of Fig. \ref{fig2} illustrate more intricate motions. For instance, in the third image of the third row, three different types of demonstrations commence from distinct initial points but all culminate at the same target point, underscoring the algorithm's versatility and effectiveness.

The quantified results are shown Table \ref{tab:table2}, the results clearly demonstrate that the proposed algorithm outperforms other methods in terms of trajectory reproduction accuracy. Showcasing substantial improvements of approximately $15.56\%$ in SEA score and approximately $13.23\%$ in $V_{rmse}$ when compared to CLF-DM.
\begin{figure}[htbp]
	\centerline{\includegraphics[width=0.32\textwidth]{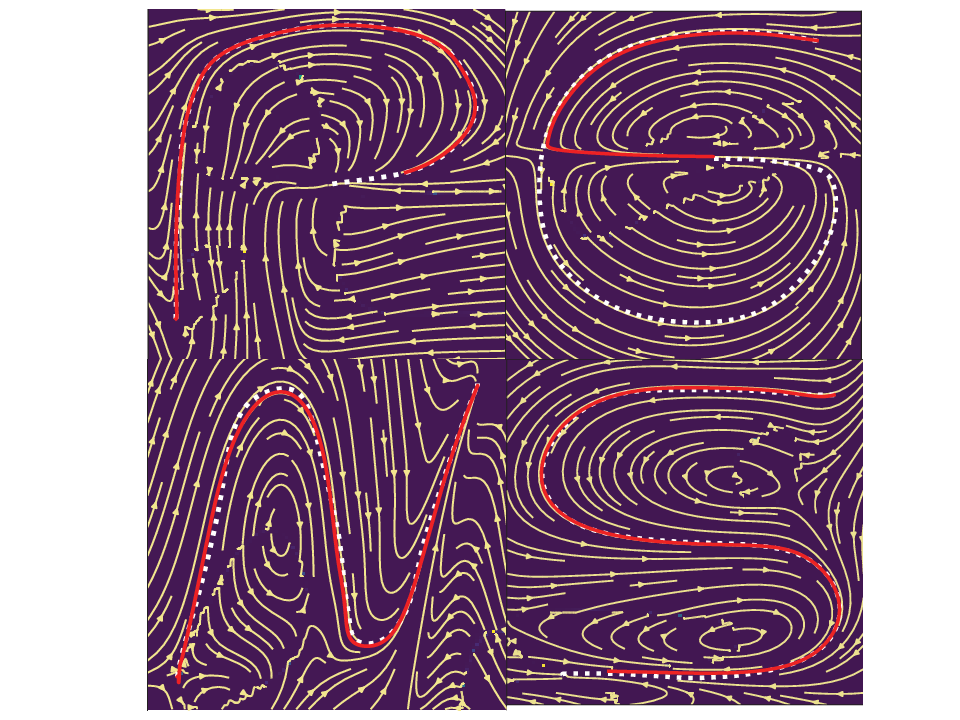}}
	\caption{Four simulation scenarios of learning from one demonstration with modeling the $\dot{\boldsymbol{y}}=-\boldsymbol{y}$  are depicted. }	
	\label{fig4}
\end{figure}
Furthermore, a noteworthy observation stems from the images presented in Fig. \ref{fig2}, characterized with white background, representing the transformed trajectories denoted as $\boldsymbol{y}$ in Fig. \ref{fig1}. To elucidate,  considering the case of the letter "J" for illustration. These transformed trajectories manage to retain certain essential characteristics of the original trajectories while scaling the two-dimensional coordinates to align with the simplest Lyapunov energy function. This innovative approach, albeit seemingly straightforward at first glance, yields intriguing results when addressing the problem at hand. 
\begin{table}[t]
	\caption{Variation in Reproduction Errors of different DS approaches on the LASA dataset\label{tab:table2}}
	\centering
	\setlength{\tabcolsep}{1.4mm}{	\begin{tabular}{ccc}
			\hline
			 methods    & Mean SEA($mm^2$)   & Mean $V_{rmse}(mm/s)$
			\\
			\hline
			\makecell*[c]{SEDS \cite{b2}} & 8.18 $\times$ 10\textasciicircum{}5 & 142.9  \\
			
			\makecell*[c]{CLF-DM \cite{b5} } & 5.72 $\times$ 10\textasciicircum{}5 & 71.8  \\			
			
			\makecell*[c]{The proposed method  }  & 4.83 $\times$ 10\textasciicircum{}5 & 62.3  \\
			\hline		
	\end{tabular}}
\end{table}

The proposed algorithm can also learn the vector field from one demonstration and four results are shown in Fig. \ref{fig3}. The results show that the proposed algorithm can learn from one demonstration well, and the vector field is similar to the vector field learned from multiply demonstrations, which is consistent with common sense.

\begin{figure}[htbp]
	\centerline{\includegraphics[width=0.41\textwidth]{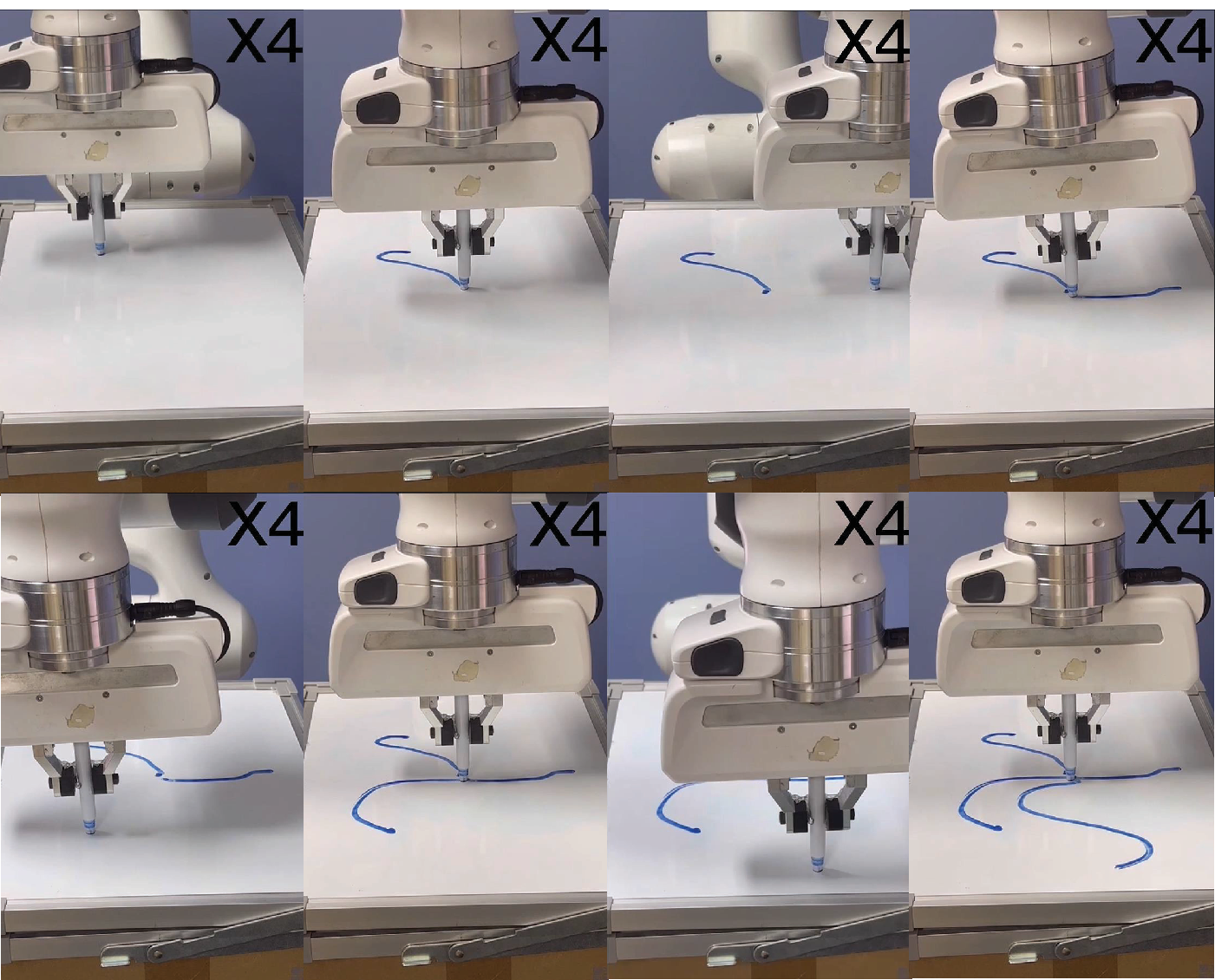}}
	\caption{The real-world experiments involving the generation of multiple trajectories using the "Multi-Models-1" dataset from LASA. These experiments employ the proposed algorithm for trajectory learning. }	
	\label{fig5}
\end{figure}

Incorporating the direct design of $\dot{\boldsymbol{y}}$ as $\dot{\boldsymbol{y}}=-{\boldsymbol{y}}$ into the modeling approach is another method of simultaneously learning the stability DS with  a Lyapunov energy function. In Fig. \ref{fig5},  the simulation results for the same data previously examined in Fig. \ref{fig3} is presented using this approach. It becomes evident that the endpoints of the reproduced trajectories for $P$ and $S$ deviate slightly from the target point. This discrepancy implies that the reproduced velocities differ significantly from the demonstrated ones. Furthermore, the reproduction of $G$ does not align with the demonstrated trajectory. Consequently, this observation suggests that when employing this method, the model's fitting capacity falls short of expectations.

\subsection{ Validation on Robot}
Experiments were conducted to validate the proposed algorithm using a Franka Emika robot. The model was trained using data from the 'Multi-Models-1' subset of the LASA dataset, then it was executed on a whiteboard using a pen held by the robot.

Throughout the experiment, the robot continuously acquired its end-effector's position and utilized the proposed algorithm to calculate the corresponding velocity.

To evaluate the algorithm's performance, four random points on the whiteboard were selected. The results demonstrated that the proposed algorithm generated trajectories that converged toward manually defined target points. These trajectories closely followed the vector field, as illustrated in the third figure of the third row in Fig. \ref{fig2}.

\section{CONCLUSIONS}\label{sec5}

This paper introduces a neural network-based algorithm specifically designed for learning from either single or multiple demonstrations.  The algorithm's performance is assessed through both simulated scenarios featuring diverse handwriting examples and practical experiments involving a physical robot. The experimental outcomes provide compelling evidence regarding the efficacy of the proposed method.

Nevertheless, it is crucial to acknowledge that the proposed algorithm relies on neural networks, and the model training process is comparatively time-consuming when compared to conventional DMPs or GMMs. Applying the $\mathrm{Relu}$ function serves to constrain the implicit velocity, thereby hindering smoothness of the vector field beyond the demonstration zone. This aspect presents an opportunity for future research, aimed at enhancing the algorithm's efficiency, generalization ability, and reducing its training time.

\addtolength{\textheight}{-12cm}   





\end{document}